# FOLIAGE PLANT RETRIEVAL USING POLAR FOURIER TRANSFORM, COLOR MOMENTS AND VEIN FEATURES

Abdul Kadir[1], Lukito Edi Nugroho[2], Adhi Susanto[3] and Paulus Insap Santosa[4]

Department of Electrical Engineering, Gadjah Mada University, Yogyakarta, Indonesia

```
1akadir@mti.ugm.ac.id
2lukito@mti.ugm.ac.id
3susanto@te.ugm.ac.id
4insap@mti.ugm.ac.id
```

## ABSTRACT

*This paper proposed a method that combines Polar Fourier Transform, color moments, and vein features to retrieve leaf images based on a leaf image. The method is very useful to help people in recognizing foliage plants. Foliage plants are plants that have various colors and unique patterns in the leaf. Therefore, the colors and its patterns are information that should be counted on in the processing of plant identification. To compare the performance of retrieving system to other result, the experiments used Flavia dataset, which is very popular in recognizing plants. The result shows that the method gave better performance than PNN, SVM, and Fourier Transform. The method was also tested using foliage plants with various colors. The accuracy was 90.80% for 50 kinds of plants.*

## KEYWORDS

*Color Moments, Plant Retrieval, PFT (Polar Fourier Transform), PNN, SVM, Vein features*

## 1. INTRODUCTION

Several researches in leaf identification have been explored, but there are still many challenges for researchers to try other approaches for better performance of the identification system. A certain method may give good performance in specific samples of leaves, but does not guarantee to perform good result for other ones. For example, ignoring colors in recognizing a foliage leaf is too risky. Sometime it is found that two or more plants have leaves with similar or same shape, but different colors. In that case, color features cannot be neglected.

Therefore, a proposed method was accomplished to retrieve plants based on a leaf query using combination features: Polar Fourier Transform (PFT), colour moments, and vein of leaf. PFT is used to handle the shape of the leaf, colour moments are used to capture color information, and features extracted from the leaf's vein are used to improve performance of the retrieval system. This approach can be used to identify a leaf and also give top five of plants that have similar properties to the leaf query.

In order to assess the performance of the system, a leaf plant dataset came from Wu et al. [1] had been used. The result shows that the combination of PFT, colour moments, and vein of leaf features gives better performance than using SVM, PNN, and Fourier Moment [2]. The same method was also tested using 50 kinds of foliage plants that contain various colors. The average accuracy is 90.80%.

The remainder is organized as follows: Section 2 discusses related works, Section 3 describes all features used in the research, Section 4 explains how the mechanism of experiments is accomplished, Section 5 presents the experimental results, and Section 6 concludes the results.

## 2. RELATED WORKS

Several researches in plant identification are described here. Warren [3] create a system that can measure the length and width of Chrysanthemum leaf and assesses several descriptive characters including the shape of the leaf apex, the shape of the base, the degree to which the margin is serrated and the depth and shape of the leaf's lower sinus. Wu et al. [4] identified 6 species of plants. No color information was processed. They used aspect ratio, leaf dent, leaf vein, and invariant moment to identify plant. Wang et al. [5] used centroid-contour distance as shape features. Du et al. [6] captured the leaf shape polygonial approximation and algorithm called MDP (modified dynamic programming) for shape matching. Wu et al. [1] proposed an algorithm for plant recognition using Probabilistic Neural Network (PNN). They used 12 geometric features of leaf as input of identification system. By using 32 kinds of plants, the system has average accuracy 90,312%. The PNN algorithm is very fast to identify a leaf. However, manual processing should be done to locate the two terminal points of the leaf. Other researches, such as Singh et al. [2] proposed Support Vector Machine (SVM) to improve the performance of the indentification system, based on data came from Wu et. al [1], and Zulkifli [7] worked on 10 kinds of leaves and uses invariants moments as features to recognize them. The previous researchers did not incorporated color features in recognizing plants, but Man et al. did [8]. Man et. al. used features came from color and texture features and utilized SVM for classification. The accuracy of their system is 92% for 24 categories.

Other features that are incorporated in recognizing plants are extracted from vein of leaf (venation). Nam et al. [9] used shape and venation for leaf image retrieval. They used MPP (Minimum Perimeter Polygons) algorithm to solve the shape of the leaf and the type of venations as venation features. Li et al. [10] used ICA (Independent Component Analysis) to extract leaf vein. Other approach has done by Wu et al. [1], by performing morphological opening on grayscale image with flat, disk-shaped structuring element of radius 1, 2, 3, and 4. Then, by subtracting remaining image by the margin, the appearance of vein like image is obtained. That operation is quite simple and fast.

Meanwhile, Polar Fourier Transform (PFT) has been introduced in [11] to recognize 52 kinds of foliage plants. Compared to other methods (moment invariants and Zernike moments), PFT is prospective for recognizing shape of plants. However, using PFT only for foliage plant retrieval is not enough. Foliage plants mean that plants have leaves with unique shape, fancy pattern, or attractive colors. Therefore, handling color for features is a must, because some leaves of different species have same patterns but different colors.

## 3. FEATURES FOR PLANT RETRIEVAL

### 3.1. Polar Fourier Transform

Fourier transform (FT) is very populer in image processing, especially for analyzing purpose. The advantage of analyzing image in spectral domain over analyzing shape in spatial domain is that it is easy to overcome the noise problem which is common to digital images [12]. However, direct applying 2-D FT on a shape image in Cartesian space to derive feature descriptors is not practical due to property of FT that is not rotation invariant. To overcome that problem, Zhang [12] derived 2 kinds of Polar Fourier Transforms (PFT). One of them is defined as follow:

$$PF2(\rho,\phi) = \sum_r \sum_i f(\rho,\phi_i) \exp[j2\pi(\frac{r}{R}\rho + \frac{2\pi}{T}\phi)] \qquad (1)$$

where

- $0 \leq r < R$ dan $\theta_i = i(2\pi/T)$ $(0 \leq i < T); 0 \leq \rho < R,\ 0 \leq \phi < T,$
- R is radial frequency resolution,
- T is angular frequency resolution.

How to compute PFT described as follow. For example, there is an image I = {f(x, y); $0 \leq x < M$, $0 \leq y < N$}. Firstly, the image is converted from Cartesian space to polar space $I_p$ = {f(r,θ); $0 \leq r < R$, $0 \leq \theta < 2\pi$ }, where R is the maximum radius from centre of the shape. The origin of polar space becomes as centre of space to get translation invariant. The centroid $(x_c, y_c)$ calculated by using formula:

$$x_c = \frac{1}{M}\sum_{x=0}^{M-1} x, y_c = \frac{1}{N}\sum_{x=0}^{N-1} y, \qquad (2)$$

In this case, (r,θ) is computed by using:

$$r = \sqrt{(x-x_c)^2 + (y-y_c)^2}, \theta = \arctan\frac{y-y_c}{x-x_c} \qquad (3)$$

Rotation invariance is achieved by ignoring the phase information in the coefficient. Consequently, only the magnitudes of coefficients are retained. Meanwhile, to get the scale invariance, the first magnitude value is normalized by the area of the circle and all the magnitude values are normalized by the magnitude of the first coefficient. So, the Fourier descriptors are:

$$FDs = \{\frac{PF(0,0)}{2\pi r^2}, \frac{PF(0,1)}{PF(0,0)}, ....,\frac{PF(0,n)}{PF(0,0)}, ....,\frac{PF(m,0)}{PF(0,0)}, ....,\frac{PF(m,n)}{PF(0,0)}\} \qquad (4)$$

where m is the maximum number of the radial frequencies and n is the maximum number of angular frequencies.

Fig. 1 shows two kinds of leaves and Fig. 2 denotes the Fourier descriptors of the two leaves. In this case, the radial frequency is equal 4 and the maximum number of angular frequencies is equal 6.

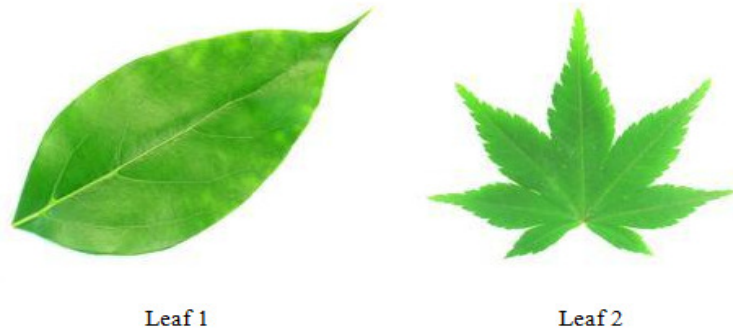

Leaf 1               Leaf 2

Figure 1.  Example of Leaves

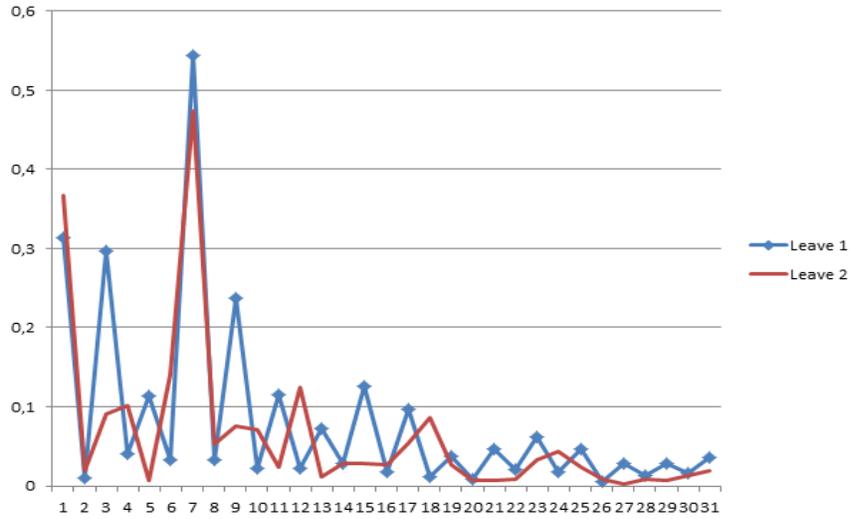

Figure 2. Example of Fourier descriptors

## 3.2. Color Moments

Color moments represent color features to characterize a color image. For example, it was used for skin texture recognition [13] and for image retrieval [14]. Features can be involved are mean (μ), standard deviation (σ), and skewness (θ). For RGB color space, the three features are extracted from each plane R, G, and B. The formulas to capture those moments:

$$\mu = \frac{1}{MN} \sum_{i=1}^{M} \sum_{j=1}^{N} P_{ij} \tag{5}$$

$$\sigma = \sqrt{\frac{1}{MN} \sum_{i=1}^{M} \sum_{j=1}^{N} (P_{ij} - \mu)^2} \tag{6}$$

$$\theta = \sqrt[3]{\frac{1}{MN} \sum_{i=1}^{M} \sum_{j=1}^{N} (P_{ij} - \mu)^3} \tag{7}$$

M and N are the dimension of image. $P_{ij}$ is values of color on column $i_{th}$ and row $j_{th}$.

## 3.3. Vein Features

Vein features are obtained by using morphological opening [1]. That operation is performed on the gray scale image with flat, disk-shaped structuring element of radius 1, 2, 3, 4 and subtracted remained image by the margin. As a result, a structure like vein is obtained. Fig. 3 shows an example of vein resulted by such operation.

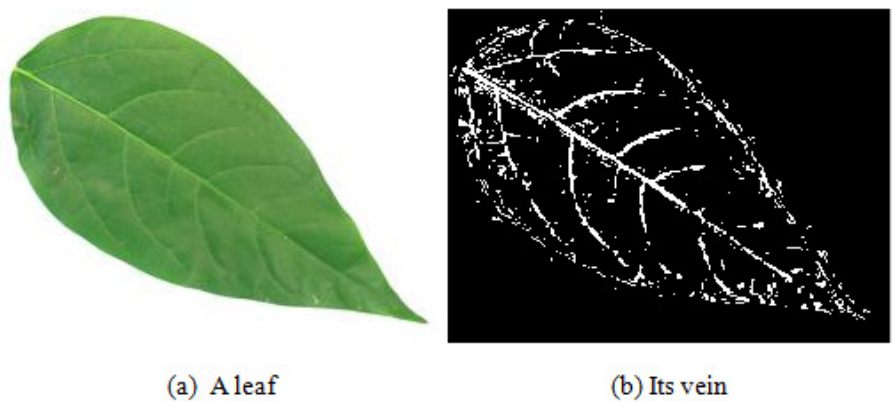

(a) A leaf         (b) Its vein

Figure 3. Illustration of vein processed by using morphological operation

Based on that vein, 4 features are calculated as follow:

$$V_1 = \frac{A_1}{A}, V_2 = \frac{A2}{A}, V_3 = \frac{A_3}{A}, V_4 = \frac{A_4}{A} \tag{8}$$

In this case, $V_i$, $V_i$, $V_i$, and $V_i$ represent features of the vein, $A_i$, $A_i$, $A_i$, and $A_i$ are total pixels of the vein, and $A$ denotes total pixels on the part of the leaf.

## 4. PROCEDURE OF LEAVES RETRIEVAL

The procedure of leaves retrieval is presented in Fig. 4. Firstly, image of the leaf as a query is processed to separate the leaf from its background. After the area of the leaf is found, all features contained in the leaf are extracted. Then, similarity indexes are calculated by comparing all features came from the leaf of the query and all features in the database. Finally, top n species are presented as a result of the query.

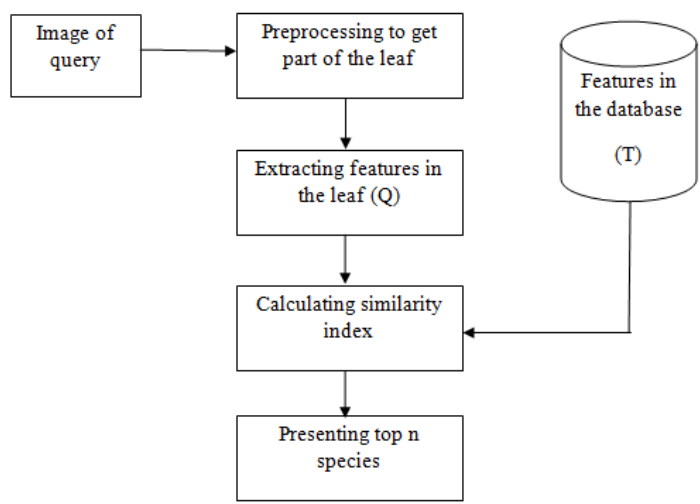

Figure 4.  Procedure of retrieving leaves

## 4.1. Processing to Get Part of the Leaf

Segmentation is a process to separate objects from its background. Several approaches have been introduced to do it. For example, the fissures of lung lobes are not seen by naked eyes in low dose Computerized Tomography images [15]. Therefore, a special treatment should be done for this case.

Separation part of the leaf is obtained by accomplishing several steps. Fig. 5 describes all processes to get part of the leaf. Conversion from RGB image to gray scale image is done by using formula below [1]:

$$gray = 0.2989 * R + 0.5870 * G + 0.1140 * B \qquad (9)$$

where R, G, B correspond to the color of pixel respectively.

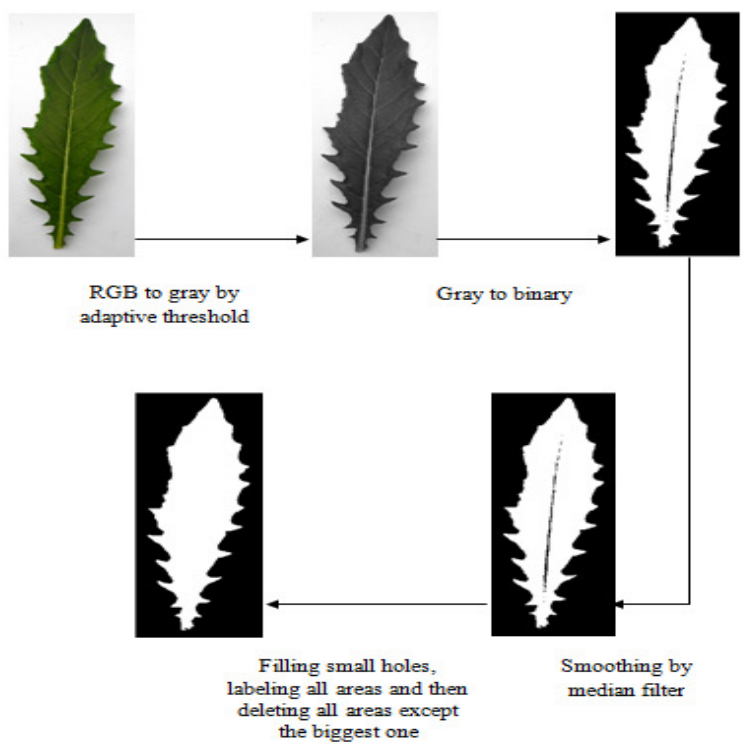

Figure 5. Schema of separating the leaf from its background

Because lightning when images were taken by camera differs from time to time, a constant threshold cannot be applied to convert the gray scale image into binary form. Therefore, adaptive threshold should be used. An algorithm to obtain such threshold uses three steps [16].

1. Construct an intensity histogram with n bin.
2. Obtain the two major peaks (range that give the maximum number of pixels) which correspond to the leaf and the background.
3. Obtain a bin between two the major peaks that give the minimum of pixels. Then, median of that bin is calculated as the threshold.

Fig. 6 shows an example of intensity histogram with 20 bins. The real values as follows: 0, 0, 322, 7696, 17057, 24685, 10739, 1638, 1533, 1651, 1982, 2329, 2582, 4010, 5574, 23558, 13704, 5143, 17886 and 24661. The two major peaks contain in bin 6 and 20. So, median of bin 10 becomes the threshold.

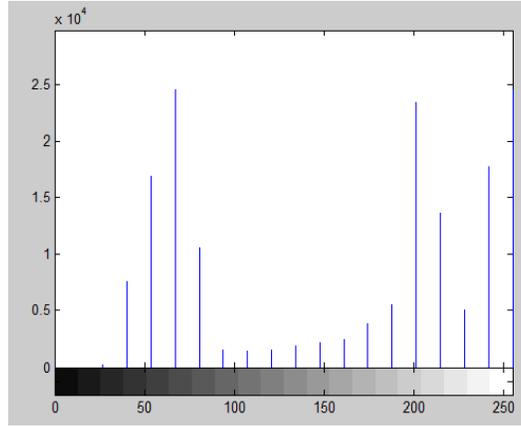

Figure 6. Intensity histogram of a leaf

A median filter is used to minimize noises contained in the image. However, the result may contain holes in the leaf due to the thresholding effect or small parts in the background caused by dirty background. To overcome those problems, holes on the leaf can be filled by using morphological operations [17]. Then, all areas in the image are labelled using 8-neighborhood. After that, all areas that have been labelled are deleted except the biggest one. As a result, part of the leaf is found.

## 4.2. Extracting Features

Features are extracted after the area of the leaf is obtained. Three kinds of features, Fourier descriptors, color moments, and vein features, are calculated based on formulas described on Section 2.

## 4.3. Calculating The Similarity Index

Similarity index of the two leaves is computed using Euclidean distance by using formulas below:

$$d_s(Q_s, T_s) = \sqrt{\sum_{i=0}^{n_s} (Q_{si} - T_{si})^2} \qquad (10)$$

$$d_c(Q_c, T_c) = \sqrt{\sum_{i=0}^{n_c} (Q_{ci} - T_{ci})^2} \qquad (11)$$

$$d_v(Q_v, T_v) = \sqrt{\sum_{i=0}^{n_v} (Q_{vi} - T_{vi})^2} \qquad (12)$$

In this case, $d_s(Q_s, T_s)$, $d_c(Q_c, T_c)$, and $d_v(Q_v, T_v)$ represent similarity index of shape, color, and vein features respectively whereas $n_s$, $n_c$, and $n_v$ represent features number of shape, color, and vein features respectively.

Similarity index that integrates Euclidean distance of shape, color, and vein is calculated using equation suggested by Jyothi et al. [18] in integrating several features:

$$d(Q,T) = \frac{w_s d_s(Q_s,T_s) + w_c d_c(Q_c,T_c) + w_v d_v(Q_v,T_v)}{w_s + w_c + w_v} \qquad (13)$$

where $w_s$, $w_c$, and $w_c$ are weights of features. In their experiment, Jyothi et. al. used all weights equal 1, but in our experiment all weights are not equal 1 to get optimum performance.

### 4.4. Presenting Top n Species

Top n species is obtained by sorting the all similarity indexes first. After that, it is easy to get n species with smallest similarity index. For achieving performance of the system, n are chosen 1, 3, and 5.

## 5. EXPERIMENTAL RESULTS

There are two kinds of dataset used in the experiments. First dataset came from Wu et. al. [1], that can be downloaded at http://flavia.sourceforge.net/. There are 32 plants. Samples leaves of each plant are shown in Fig. 7. Second dataset came from our collection (called Folia dataset, that can be downloaded at http://www.mediafire.com/?k9cx9ro4c352baf and http://www.mediafire.com/?qmawxsiv3kge2d1). The dataset contains 50 kinds of leaves (48 kinds of plants), as shown in Fig. 8. Actually, there are two plants that are represented by 2 kinds of leaves (denoted by flags * and ** in Fig. 8).

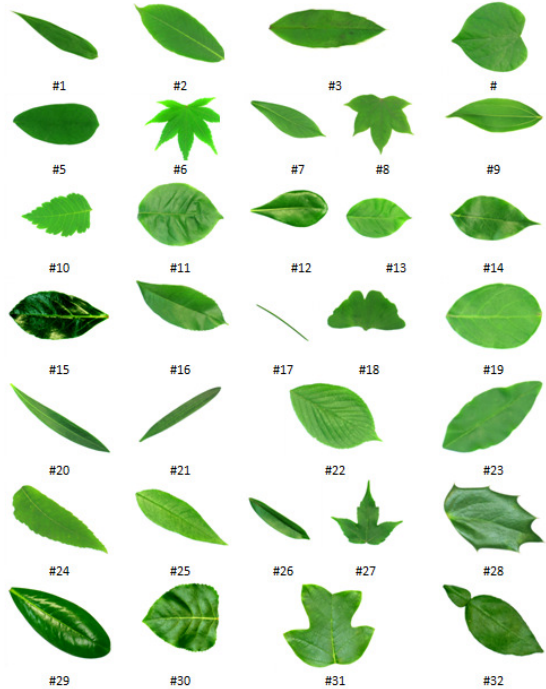

Figure 7. Leaves in Flavia dataset

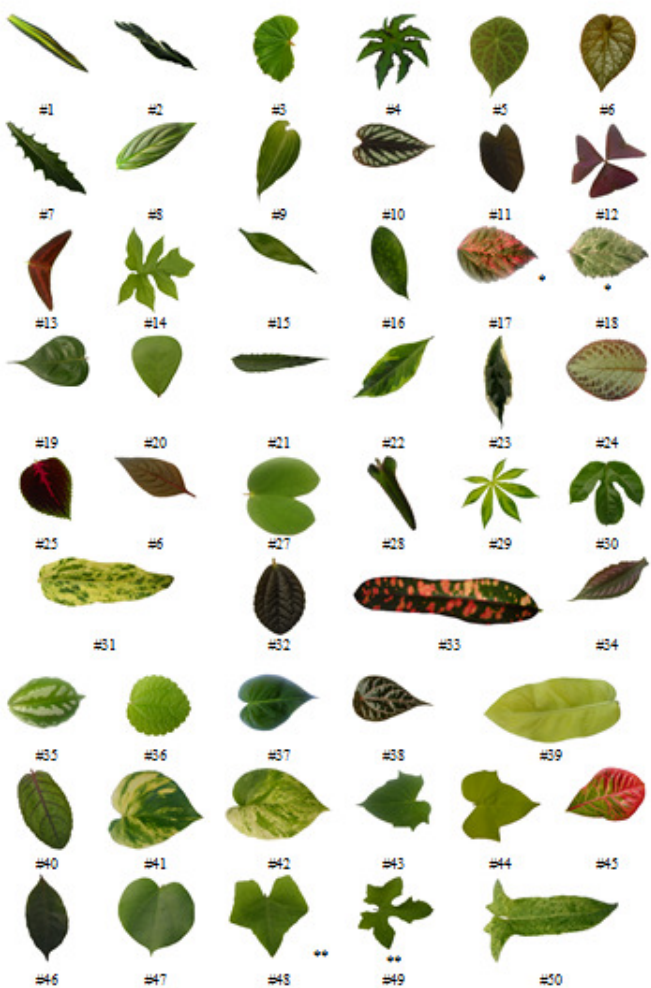

Figure 8. Leaves in Folia dataset

In order to obtain performance of the system, we used the following formula [18]:

$$Performance = \frac{Number\ relevant\ of\ images}{Total\ number\ of\ query} \qquad (14)$$

Based on Eq. 13, we set $w_s$ = 0.55, $w_c$ = 0.25, and $w_v$ = 0.20 to test Flavia and Folia databases. By using 40 leaves per species as references and 10 leaves per species as testing leaves, the system performance using Flavia dataset reaches 93.13% for top 1, 97.81% for top 3, and 98.12% for top 5. In this case, the radial frequency for calculating PFT is equal 4 and the maximum number of angular frequencies is equal 6.

Table 1 shows the accuracy of our methods compared to other methods per species. Integration of Fourier descriptor, color moment, and vein features has better performance than PNN, SVM, and Fourier moments.

Table 1.  Comparison accuracy of several methods using Flavia dataset

| Species | SVM [1] | PNN-PCNN [1] | Fourier Moment [1] | PFT+ |
|---------|---------|--------------|--------------------|------|
| #1 | 100% | 100% | 32.2% | 90% |
| #2 | 100% | 90% | 0% | 100% |
| #3 | 100% | 100% | 79.17% | 80% |
| #4 | 100% | 85% | 63.01% | 100% |
| #5 | 100% | 100% | 30.36% | 100% |
| #6 | 80% | 85% | 52.9% | 100% |
| #7 | 100% | 100% | 92.31% | 100% |
| #8 | 90% | 90% | 33.9% | 100% |
| #9 | 80% | 95% | 0% | 90% |
| #10 | 80% | 90% | 29.23% | 100% |
| #11 | 90% | 95% | 28% | 100% |
| #12 | 0% | 90% | 69.84% | 100% |
| #13 | 90% | 90% | 25% | 100% |
| #14 | 100% | 85% | 1.54% | 100% |
| #15 | 0% | 90% | 68.33% | 100% |
| #16 | 80% | 75% | 48.21% | 100% |
| #17 | 100% | 100% | 97.4% | 100% |
| #18 | 100% | 95% | 85.48% | 100% |
| #19 | 80% | 90% | 81.97% | 70% |
| #20 | 100% | 85% | 93.94% | 100% |
| #21 | 90% | 80% | 90% | 100% |
| #22 | 90% | 85% | 27.27% | 10% |
| #23 | 80% | 90% | 0% | 100% |
| #24 | 80% | 95% | 56.92% | 90% |
| #25 | 40% | 85% | 1.85% | 100% |
| #26 | 90% | 85% | 75% | 80% |
| #27 | 100% | 100% | 95.35% | 90% |
| #28 | 100% | 100% | 5.45% | 90% |
| #29 | 90% | 80% | 15.79% | 100% |
| #30 | 0% | 100% | 0% | 100% |
| #31 | 80% | 100% | 20.77% | 100% |
| #32 | 100% | 90% | 80.36% | 90% |
| Average | 81,56% | 91.25% | 46.30% | 93.13% |

Table 2 shows the results using Flavia and Folia datasets. Although the performance of Folia testing is lower than such in Flavia dataset, combination of PFT, color moments, and vein features is prospective for foliage plant identification, because they can capture colors in the leaf. For example, it is very useful to differ two plants such as *Epipremnum pinnatum* 'Aureum' (#41 in Fig. 8) and *Epipremnum pinnatum* 'Marble Queen' (#42 In. Fig. 8) that have similar patterns but different colors.

Table 2.  Performance of the system using Flavia and Folia datasets

| Dataset | Top 1 | Top 3 | Top 5 |
|---------|-------|-------|-------|
| Flavia | 93.13% | 97.81% | 98.12% |
| Folia | 90.80% | 97.80% | 98.80% |

Table 3 shows performance of each plant when using Folia dataset. 27 of 50 plants can be identified with accuracy 100%. However, accuracy of plant such as #9 is very low.

Table 3.  Performance of the system using Folia dataset

| Species | Top 1 | Species | Top 1 |
|---------|-------|---------|-------|
| #1 | 90% | #26 | 90% |
| #2 | 70% | #27 | 100% |
| #3 | 100% | #28 | 100% |
| #4 | 100% | #29 | 70% |
| #5 | 100% | #30 | 100% |
| #6 | 80% | #31 | 100% |
| #7 | 90% | #32 | 100% |
| #8 | 80% | #33 | 100% |
| #9 | 50% | #34 | 100% |
| #10 | 90% | #35 | 100% |
| #11 | 100% | #36 | 100% |
| #12 | 90% | #37 | 100% |
| #13 | 100% | #38 | 100% |
| #14 | 80% | #39 | 100% |
| #15 | 100% | #40 | 50% |
| #16 | 70% | #41 | 90% |
| #17 | 100% | #42 | 80% |
| #18 | 100% | #43 | 90% |
| #19 | 100% | #44 | 100% |
| #20 | 100% | #45 | 90% |
| #21 | 100% | #46 | 100% |
| #22 | 100% | #47 | 90% |
| #23 | 90% | #48 | 100% |
| #24 | 60% | #49 | 80% |
| #25 | 100% | #50 | 70% |

# 6. CONCLUSIONS

A system for retrieving plants using PFT, color moments, and vein features had been developed. The system has been tested using two kinds of dataset: Flavia and Folia datasets. The system has accuracy of 93.13% when it was tested using Flavia dataset and has accuracy of 90.80% when it was tested using Folia dataset. The last accuracy shows that the integration of PFT, color moments, and vein features are good enough to be used in foliage plant retrieval. However, to improve the performance of the system, other features such as texture features can be used for future researches.

## Authors

**Abdul Kadir** received B.Sc in Electrical Engineering from Gadjah Mada University in 1987, M. Eng in Electrical Engineering from Gadjah Mada in 1998, and Master of Management from Gadjah Mada University in 2004. His research interests include image processing, pattern recognition, and web-based applications.

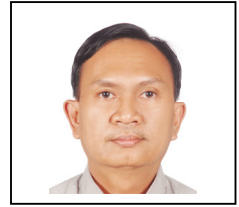

**Lukito Edi Nugroho** received B.Sc in Electrical Engineering from Gadjah Mada University in 1989, M.Sc. from James Cook University of North Queensland in 1994, Ph.D from School of Computer Science and Software Engineering, Monash University, in 2002. His research interests are software engineering, information systems, and multimedia.

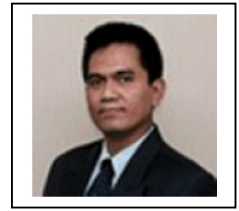

**Adhi Susanto** is a professor emeritus at Gadjah Mada University, Indonesia. He received Bachelor in Physics in 1964 from Gadjah Mada University, Master in Electrical Engineering in 1966 from University of California, Davis, USA, and Doctor of Philosophy in 1986 from University of California, Davis, USA. His research interests areas are electronics engineering, signal processing, and image processing.

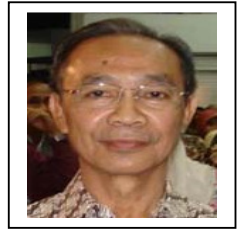

**Paulus Insap Santosa** obtained his undergraduate degree from Universitas Gadjah Mada in 1984, master degree from University of Colorado at Boulder in 1991, and doctorate degree from National University of Singapore in 2006. His research interests include human computer interaction and technology in Education.

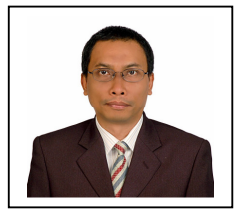